\pdfoutput=1

\documentclass[11pt]{article}

\usepackage{acl}

\usepackage{times}
\usepackage{latexsym}
\usepackage{graphicx}
\usepackage{float}
\usepackage{color, colortbl}
\definecolor{Gray}{gray}{0.9}
\usepackage{graphicx}
\usepackage{makecell}
\usepackage{lipsum}
\usepackage{float}
\usepackage{placeins}
\usepackage{hyperref}
\usepackage{epstopdf}
\usepackage{algorithm}
\usepackage{amsmath}
\usepackage{amssymb}
\usepackage{multirow}
\usepackage{subcaption}

\usepackage[T1]{fontenc}

\usepackage[utf8]{inputenc}

\usepackage{microtype}

\title{Meme-ingful Analysis: Enhanced Understanding of
Cyberbullying in Memes Through Multimodal Explanations}

\author{Prince Jha$^{1^\star}$,
Krishanu Maity$^{1^\star}$,
Raghav Jain$^{1^\star}$,
Apoorv Verma$^{1}$,
Sriparna Saha$^{1}$ \and \\ 
\textbf{Pushpak	Bhattacharyya}$^{2}$\\
$^{1}$Department of Computer Science and Engineering, Indian Institute of Technology Patna\\ $^{2}$ Department of Computer Science and Engineering, Indian Institute of Technology Bombay\\
\texttt{princekumar\_1901cs42@iitp.ac.in}
\thanks{\textsuperscript{*} The first three authors contributed equally to this work and are jointly the first authors.}}

\begin{document}
\maketitle
\begin{abstract}
Internet memes have gained significant influence in communicating political, psychological, and sociocultural ideas. While memes are often humorous, there has been a rise in the use of memes for trolling and cyberbullying. Although a wide variety of effective deep learning-based models have been developed for detecting offensive multimodal memes, only a few works have been done on explainability aspect. Recent laws like "right to explanations" of General Data Protection Regulation, have spurred research in developing interpretable models rather than only focusing on performance. Motivated by this, we introduce {\em MultiBully-Ex}, the first benchmark dataset for multimodal explanation from code-mixed cyberbullying memes. Here, both visual and textual modalities are highlighted to explain why a given meme is cyberbullying. A Contrastive Language-Image Pretraining (CLIP) projection-based multimodal shared-private multitask approach has been proposed for visual and textual explanation of a meme. Experimental results demonstrate that training with multimodal explanations improves performance in generating textual justifications and more accurately identifying the visual evidence supporting a decision with reliable performance improvements.\footnote{\url{https://github.com/Jhaprince/MemeExplanation}}\\
{\bf Disclaimer:} The article contains profanity, necessary for the nature of the work, but not reflecting the authors' opinions.
\end{abstract}

\section{Introduction}
\begin{figure}[hbt]
	\centering
	\includegraphics[height = 5.5 cm, width = 7.5cm]{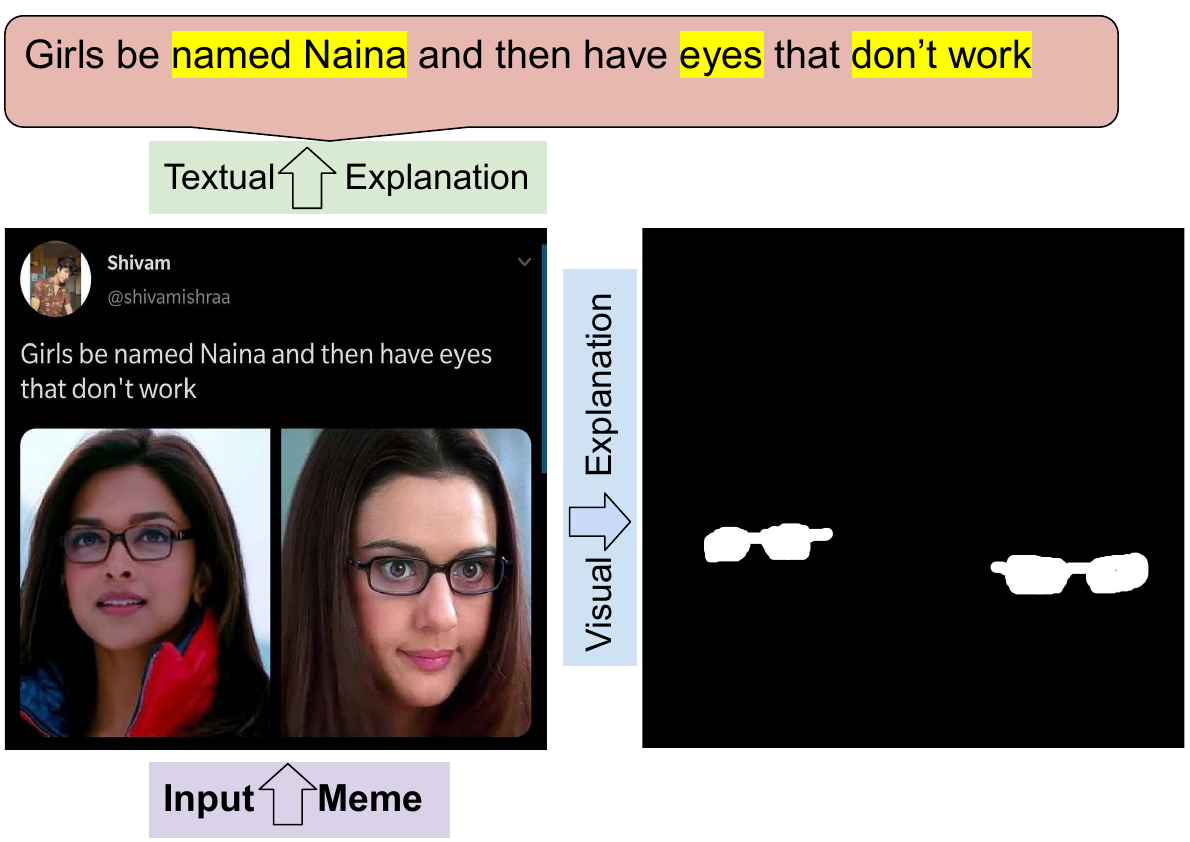}
	\caption {Cyberbullying Explanation in memes. Here the aim is to highlight both the image and text as an explanation of why the given meme is a bully.}
	\label{fig:samp}
\end{figure} 
The tremendous increase in multimodal content due to the widespread use of social media platforms renders human moderation of such information untenable~\cite{cao2020deephate}. Memes, which are images with tiny text descriptions embedded in them, have become a popular kind of multimodal content on social media in recent years. 
Though memes
are typically humorous, it also stimulates the propagation of online abuse and harassment, including cyberbullying. Cyberbullying~\cite{smith2008cyberbullying} is any communication that disparages an individual on the basis of a characteristic such as color, gender, race, sexual orientation, ethnicity, nationality, or other features. The Pew Research Center estimates that 40\% of social media users have encountered online harassment or bullying\footnote{\url{https://www.pewresearch.org/internet/2017/07/11/online-harassment-2017/}}~\cite{chan2019cyberbullying}. Cyberbullying victims may experience despair, worry, low self-esteem, and even suicidal thoughts~\cite{sticca2013longitudinal}.
Automatic cyberbullying detection techniques with the model's explainability are highly required to minimize those unpleasant consequences.\\
\textbf{Motivation and Evidence:} Over the last decade, studies on cyberbullying detection have focused primarily on textual content~\cite{agrawal2018deep,dadvar2014experts,paul2020cyberbert} and, recently on memes~\cite{DBLP:conf/nips/KielaFMGSRT20,DBLP:conf/emnlp/PramanickSDAN021,DBLP:conf/sigir/MaityJ0B22} in monolingual setting, with limited research focusing on code-mixed language. The use of code-mixed languages in different social media and message-sharing apps proliferates rapidly in multilingual countries~\cite{rijhwani2017estimating}. Code Mixing is a  linguistic phenomenon where words or phrases from one language are inserted into an utterance from another language~\cite{cm1}.
However, those researchers mostly concentrated on improving the performance of detecting offensive posts using various deep learning models without giving any insight or analysis into the explainability. 
Consequently, we propose a novel problem called \textbf{ Multimodal Explanation of Code-Mixed Cyberbullying Memes (MExCCM)}. This task involves processing multimodal inputs and aims to generate both textual and visual explanations for multimodal cyberbullying memes.\\
\textbf{Research Gap}: Till now, most of the works on offensive memes are limited to classification tasks. In the explainability aspect, there are some works on text data only (highlighting the words or phrases in a sentence)~\cite{mathew2020hatexplain, karim2021deephateexplainer} and only one work on multimodal memes (internal layers' attention weight visualization)~\cite{DBLP:conf/www/HeeLC22}. Still, there is no work where both text and images are highlighted to justify the offensiveness of cyberbullying content like a human does. 
Thus, to mitigate the above-mentioned research gap, we aim to build a deep learning-based model that can explain cyberbullying nature of memes in both visual and textual modalities. We seek this idea from semiotic textology linguistic theory~\cite{garcia2020borreguero}, which includes three subcomponents in order to consider how each textual media derives meaning; \textit{dictum} (aka denotation), \textit{evocatum} (aka connotation), and \textit{apperceptum} (mental images), the latter one embodying the vision-grounded analysis of textual content.\\
\textbf{Contributions}: Our contributions are threefold:
(i) We present \textit{MExCCM}, a novel task for generating multimodal explanations for code-mixed cyberbullying memes, a first in this field.
(ii) We introduce \textit{MultiBully-Ex}, the first multimodal explainable code-mixed cyberbullying dataset. It includes manual highlighting of both text and image modalities in a meme to demonstrate why it is considered bullying (iii) We propose an end-to-end Contrastive Language-Image Pretraining (CLIP) approach for visual and textual meme explanation, aiming to encourage more research on code-mixed data.
\section{Related Works}
Cyberbullying is very reliant on linguistic subtlety. Researchers have recently provided a lot of attention to automatically identifying cyberbullying in social media. In this section, we will review recent works on the detection and explainability aspects of cyberbullying.\\
\textbf{Detection:} Researchers have made significant strides in detecting meme-based cyberbullying and offensive content. \citet{DBLP:conf/sigir/MaityJ0B22} created MultiBully, a Twitter and Reddit dataset in code-mixed language, proposing two multitask platforms for detecting bullying memes, sentiment, and emotion. \citet{DBLP:conf/emnlp/PramanickSDAN021} extended the HarMeme dataset and developed a deep multimodal network to detect harmful memes, focusing on COVID-19 and US politics. Other notable works include \citet{DBLP:conf/nips/KielaFMGSRT20}'s hate speech detection with 69.47\% accuracy using Visual-BERT, \citet{gomez2020exploring}'s MMHS150K dataset of 150K Tweets, and \citet{suryawanshi2020multimodal}'s MultiOFF dataset for identifying offensive meme content, which showcased a fusion method combining text and image modalities.\\
 \textbf{Explainability:} LIME~\cite{ribeiro2016should} and SHAP~\cite{lundberg2017unified} have been used to advance both textual and visual explainability in machine learning models. \citet{zaidan2007using} improved sentiment classification by employing human-annotated "rationales." \citet{mathew2020hatexplain} introduced \textit{HateXplain}, finding that training with human rationales reduced bias. \citet{karim2021deephateexplainer} developed an explainable hate speech detection in Bengali, highlighting crucial words. \citet{DBLP:conf/www/HeeLC22} visualized how VilBERT and VisualBERT models captured slurs in hateful memes, discovering the image modality's significant contribution. While most studies used explainability to justify model outputs, our task uniquely focuses on explainability as the output itself, specifically designed to offer textual and visual explanations for cyberbullying memes. This represents the first effort to generate \textit{MExCCM}.

\section{Multimodal Bully Explanations Dataset {\em (MultiBully-Ex)}}
To create {\em MultiBully-Ex}, we utilize {\em MultiBully } dataset\footnote{\url{https://github.com/Jhaprince/MultiBully}}~\cite{maity2022multitask}, which includes 3222 bully and 2632 nonbully memes. We selected this dataset because it is the only openly available meme dataset on cyberbullying in a code-mixed setting. Our work focuses on jointly extracting textual rationales (words or phrases) and visual masks (image segmentation) to localize salient regions to explain cyberbullying detection tasks. Hence we only considered the bully memes for further annotation.
\subsection{Annotation training}
The annotation was led by three Ph.D. scholars with adequate knowledge and expertise in detection and mitigation of cyberbullying, hate speech, and offensive content  and performed by three undergraduate students with proficiency in both Hindi and English. First, ten undergraduate computer science students were voluntarily hired through the department email list and compensated through honorarium\footnote{refer to Appendix~\ref{sec:days} and Appendix~\ref{sec:cost} for more details on cost and timeline}. For annotation training, we required gold standard samples annotated with rationale labels. We aim to annotate the text explainability (rationales) part first, and then, based on those rationales, the visual annotation will be done. Our expert annotators randomly selected 150 memes and highlighted the words (rationales) for the textual explanation. Each word in a meme has been assigned a value of 0 or 1, where 1 represents that it is one of the rationales. Later expert annotators discussed each other and resolved the differences to create 150 gold standard samples with rationale annotations. We divide these 150 annotated examples into three sets, 50 rationale annotations each,  to carry out three-phase training. After the completion of every phase, expert annotators met with novice annotators to correct the wrong annotations, and simultaneously annotation guidelines (refer Appendix~\ref{sec:ann-guide}) were also renewed. After completing the third round of training, the top three annotators were selected to annotate the entire dataset. 
\subsection{Main Annotation}
We used the open-source platform Docanno\footnote{\url{https://github.com/doccano/doccano}} deployed on a Heroku instance for main annotation where each qualified annotator was provided with a secure account to annotate and track their progress exclusively.
 We initiated our main annotation process with a small batch of 100 memes and later raised it to 500 memes as the annotators became well-experienced with the tasks. We tried to maintain the annotators’ agreement by correcting some errors they made in the previous batch. On completion of each set of annotations, final rationale labels were decided by the majority voting method. If the selections of three annotators vary, we enlist the help of an expert annotator to break the tie. We also directed annotators to annotate the posts without regard for any particular demography, religion, or other factors.  We use the Fleiss' Kappa score~\cite{fleiss1971measuring} to calculate the token level inter-annotator agreement (IAA) among multiple raters for the rationale detection task signifying the dataset being of acceptable quality. IAA obtained a score of 0.72 for the rationales detection task signifying the dataset being of acceptable quality.
 \\
 Once annotators finished doing rationale annotations, they were further asked to highlight the visual regions that could justify the rationale annotations. Visual annotations were done using open source image segmentation UI interface label studio\footnote{\url{https://labelstud.io/}}, where the annotator has to mark the regions of the image to generate a binary image where the highlighted portion having pixel value 1 and others are 0. Figure~\ref{fig:samp} shows an annotated sample from the {\em MultiBully-Ex} dataset. We assessed the inter-annotator agreement for visual annotation using the Dice coefficient, which is a measure of overlap between two annotations. To ensure the accuracy of the annotations, we first had a single annotator create them and then assigned the same annotation to another annotator. We then compared their annotations using the Dice coefficient. If the coefficient was greater than 0.5, we included the annotation from the first annotator. However, if the coefficient was less than or equal to 0.5, an expert annotator was consulted to make the annotation. It's noteworthy that the average number of tokens highlighted as 'bully' was 6.79. Conversely, the total average number of tokens for 'meme' amounted to 14.12. Additionally, we discovered that the total average percentage of the area covered by visual explanations within the meme was 35.18\footnote{refer Appendix~\ref{sec:stats} for more details on dataset statistics}.

\section{Methodology}
\begin{figure*}[hbt]
	\centering
	\includegraphics[height = 10 cm, width = 12cm]{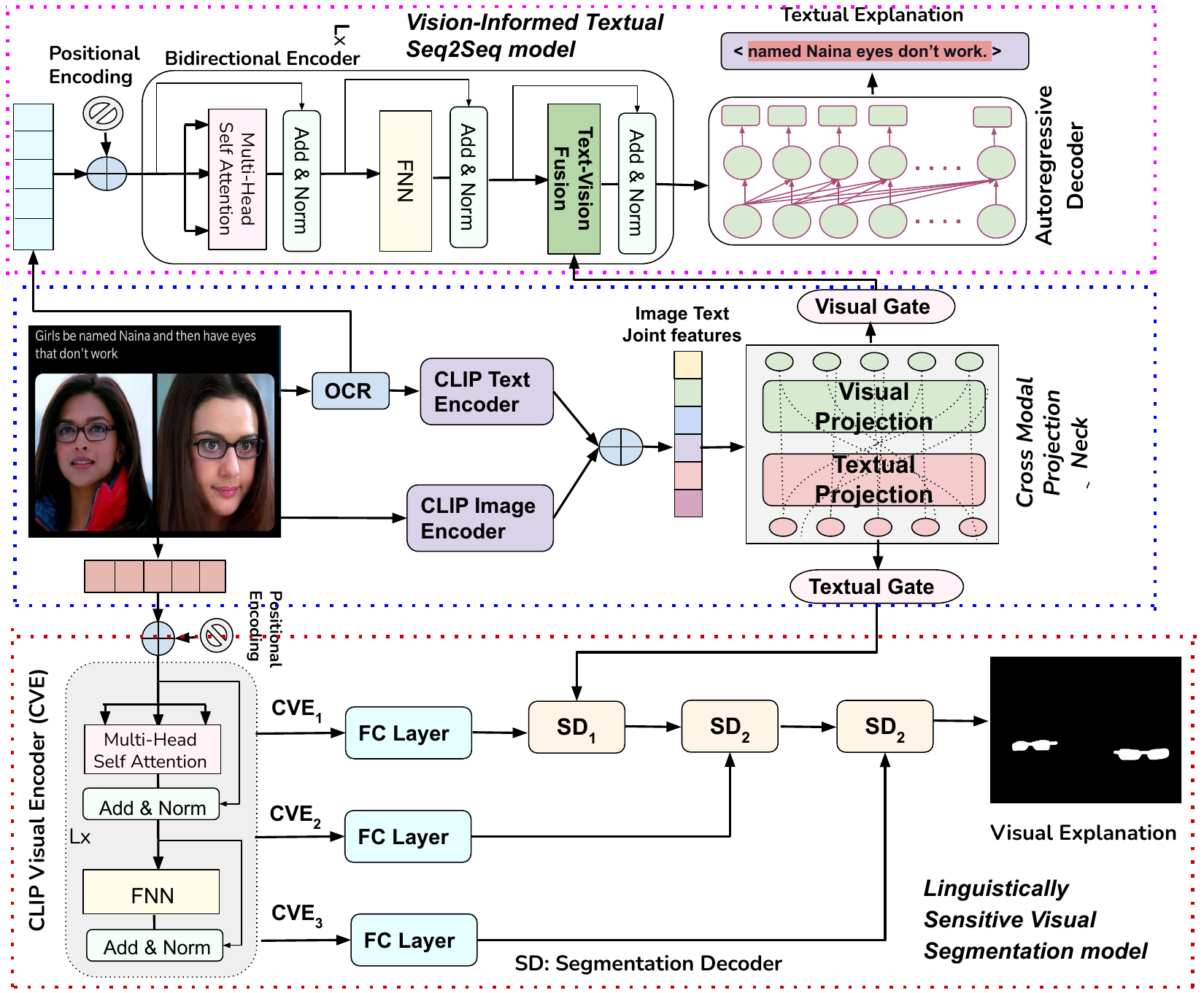}
	\caption {CLIP projection-based (CP) multimodal shared-private multitask architecture. The \textit{Vision-Informed Textual Seq2Seq model} is represented by a pink dotted box. \textit{The Cross Modal Projection Neck} is signified by a blue dotted box. The \textit{Linguistically Sensitive Visual Segmentation model }is indicated by a red dotted box. Lx denotes number of transformer layers}
	\label{fig:archi}
\end{figure*} 
\textbf{Formulation of MExCM: }Formally, given a meme ($M$) with textual modality $T = \{t_{1},t_{2},...,t_{n}\}$ and visual modality $V \in R^{3\times W\times H}$, where $W$ is the width and $H$ is the height of an image, we intend to learn textual justification along with visual evidence which is defined as follow: \textbf{(1) Textual Explanation:} Textual explanation is the process of extracting pertinent rationales $R = \{r_{1},r_{2},..,r_{k}\}$ from the textual modality $T$ of a meme $M$, which contributes to its classification as a cyberbullying instance. \textbf{(2) Visual Explanation:} Visual explanation involves a semantic segmentation task, the aim of which is to predict the segmented region $S \in R^{1\times W\times H}$ within the visual modality $V$. This segmented region is perceived as supporting evidence aligning with the textual justification. \\
Motivated from~\citet{liu2016recurrent}, we propose a CLIP~\cite{radford2021learning} projection-based (CP) multimodal shared-private multitask architecture. To enhance comprehension of our proposed method, we partition it into three distinct components: (1) CLIP Projection-Based Cross-Modal Neck, (2) Vision-Informed Textual Seq2Seq model, and (3) Linguistically-Sensitive Visual Segmentation model. In our design, the CLIP projection-based cross-modal neck acts as a shared layer, serving both the textual and visual explanation components. Meanwhile, we employ BART encoder and CVE (CLIP Visual Encoder) as private layers, enabling them to focus more effectively on their respective tasks. This decision to use separate task-specific encoders stems from our concern that a unified encoder's shared feature space might inadvertently contain task-specific features, potentially leading to unnecessary feature redundancy and a mixing of sharable features in the private space

\subsection{CLIP Projection-Based Cross-Modal Neck}
Our proposed CLIP projection-based Cross-Modal Neck acts as a common component bridging two task-specific networks: (1) the Vision-Informed Textual Seq2Seq model, and (2) the Linguistically-Sensitive Visual Segmentation model. We implement modality-specific gating mechanisms to manage the interplay of information between these textual and visual facets. The initial step in our process involves the acquisition of representations for each text-meme pair. This is facilitated by CLIP (Contrastive Language-Image Pre-training), a pre-trained model proficient in visual-linguistic tasks, which leverages its capabilities to encapsulate the holistic meaning of the meme. CLIP's effectiveness can be traced back to its extensive pre-training on 400 million image-text pairs harvested from the Internet. This training process, driven by contrastive learning objectives, along with the breadth of imagery and natural language exposure, bestows upon CLIP's remarkable zero-shot performance. In this study, we use multilingual BERT for text encoding and the Vision Transformer for image encoding. We extract two core features from each meme: a CLIP visual feature, $C_{I}$, from the meme's image, $M$, and a CLIP textual feature, $C_{T}$, from its OCR-extracted text, $T$. Both these features, $C_{I}$ and $C_{T}$, are represented as 512-dimensional vectors. After this, these two vectors ($C_{I}$ and $C_{T}$) are concatenated to create a joint vector representation of both modalities which are fed into the following two gating mechanisms simultaneously: 
\subsubsection{Gated Visual Projection}
Previous research \cite{zhang2018adaptive,lu2018visual} highlights the infeasibility of correlating functional words, such as 'the,' 'of,' and 'well,' with any visual block. To address this, our approach includes a visual gate designed to dynamically calibrate the contribution of visual features. We also employ a cross-modal projection neck to transpose gated visual features into the space of a BART (or T5) encoder. The implementation of the cross-modal projection neck can be achieved via a transformer-based architecture, leveraging its capacity to enable global attention among input tokens. To facilitate this, we feed the visual encoding from CLIP into the transformer-based network, merging it with randomly initialized, learnable weights (RW). The integration of these learnable weights serves dual purposes. Firstly, it empowers the multi-head attention mechanism with access to valuable information from the CLIP embedding. Secondly, it enables the network parameters to adapt responsively to incoming information, thereby enhancing the system's ability to learn and evolve over time.
\subsubsection{Gated Textual Projection}
Recent literature \cite{garcia2020borreguero,jha2022combining} illustrates that several communicative aspects, including facial expressions, gestures, postures, spatial relationships, color schemes, and movement, are more accurately expressed via visual cues as compared to text-based communication. In response to these findings, our proposed model incorporates a textual gating mechanism that moderates the influence of textual features. Complementing this, we utilize a Feed Forward Network (FFN) to map these textual characteristics into the domain of the segmentation decoder. This integrated approach underscores the importance of both visual and textual elements, aligning with our overarching aim of developing a multimodal understanding of memes.

\subsection{Vision-Informed Textual Seq2Seq Model}
We introduce a module designed to generate explainable text, which harnesses visual understanding by employing a combination of CLIP-based gated visual projection and generative pre-trained language models (GPLMs), specifically BART and T5. The process begins with the tokenization of input text and its transformation into a sequence of embeddings, $X_{t} \in R^{N\times d_{t}}$, where $N$ is the sequence length and $d_{t}$ is the feature dimension. To preserve the positional information of these token embeddings, positional encodings, $E_{pos_{t}} \in R^{N\times d_{t}}$ are added elementwise. The resultant input $Z_{0}$, now encompassing the positional information, is channeled into our proposed vision-aware encoder.

This vision-aware encoder comprises three subcomponents: 1) Multi-head Self-Attention (MSA), 2) Feedforward Network (FNN), and 3) Text-Vision Fusion (TVF). Additionally, each sublayer is followed by a residual connection~\cite{he2016deep} and layer normalization~\cite{ba2016layer}. The MSA (Multi-head Self-Attention) and FNN (Feedforward Network) components of our model are standard transformer layers, designed to facilitate the processing of our input data.

CLIP visual features, $C_{I}$, and textual features, $C_{T}$, are processed through the Gated Visual Projection ($GVP$) (as defined in the previous section) to yield a controlled visual information $P_{v} \in R^{M\times d_{t}}$, where $M$ is the projected sequence length with an embedding dimension of $d_{t}$.

\begin{equation}
P_{v} = GVP(C_{I},C_{T})
\end{equation}

In the Text-Vision Fusion (TVF) component of our model, we employ two types of multimodal fusion mechanisms (refer Appendix~\ref{sec:A}), namely dot product attention-based fusion and multi-head attention-based fusion as suggested in \cite{yu2021vision, tsai2019multimodal}. Formally, textual input $Z_{t} \in R^{N\times d_{t}}$ and gated visual input $P_{v} \in R^{M\times d_{t}}$ are fused to produce a vision-aware textual representation $F \in R^{N\times d_{t}}$ that has a same dimension as the textual input, which allows the continual stacking of layers.

\subsection{Linguistically Sensitive Visual Segmentation Model}
We introduce a transformer-based encoder-decoder model, inspired by the UNet architecture, that incorporates a novel gated textual projection mechanism (CP-UNet). This mechanism is designed to augment the representation capabilities of the encoder, thereby enhancing the overall efficacy of the model. Our encoder assembly includes a series of transformer layers based on the CLIP model, linked to the decoder via residual connections. The decoder is structured around a straightforward transformer-based architecture, leveraging the insights offered by the encoder to generate the final output. Formally, an input image $V \in R^{C\times W\times H}$ is processed through the CLIP visual encoder, resulting in a sequence of embeddings $X_{v} \in R^{P\times d_{t}}$, where $P$ represents the projected sequence length with dimension $d_{t}$. To encapsulate spatial features from the visual information, we incorporate a positional embedding $E_{pos_{v}} \in R^{P\times d_{t}}$. The encoded representation is acquired by passing the input through a cascade of sub-layers, including MSA and FNN, succeeded by Layer Normalization. At each layer of the CLIP visual transformer, these encodings are captured and projected into the decoder's space. They are subsequently merged with the internal features of our decoder preceding each transformer block. The decoder is designed to match the number of transformer blocks extracted from the CLIP visual transformer. Importantly, the decoder inputs are modulated with a projected gated textual vector, facilitating a deeper comprehension of the linguistic context embedded in the input, thereby yielding more accurate and contextually aligned outputs.
\subsection{Loss Prioritization}
Inspired by \citet{bengio2009curriculum}, we introduce the concept of loss prioritization sequentially so that we can concentrate on a specific task on a priority basis. The basic hypothesis is that the cognitive process of \textit{MExCCM} may not be entirely simultaneous. Both generation loss and segmentation loss must combine sequentially to achieve the desired output. We combine the loss function with a certain periodicity, i.e., after a given number of epochs $ep \in \{15,20,25\}$. The network initially learns its weight over a particular loss function (learning particular aspects of tasks), after which it self-tunes the weights over all loss functions combined sequentially (learning some other facets of the task). Mathematically, an overall global loss function, $L^{ep}_{global}$ can be defined by the equation: $    L_{global}^{ep} =  L_{i_{0}}^{0.ep} + L_{i_{1}}^{1.ep} $ where $L_{i_{q}}$s are individual losses such that $i_{q} \in$ $\{generation\_loss, segmentation\_loss\}$ and $q$ can be non-negative integer, at a given periodicity of {\em ep} epochs. A regular cross-entropy loss is employed to calculate generation\_loss and segmentation\_loss.

\section{Results and Discussion}
For a fair comparison with proposed models, we have set up standard baselines such as BART \cite{lewis2019bart}, T5 \cite{raffel2020exploring}, VG-BART, VG-T5 \cite{yu2021vision}, and DeepLabv3 \cite{chen2017rethinking}, MobileNetv3 \cite{howard2019searching}, Fully Convolutional Networks (FCN) \cite{long2015fully}, UNet \cite{ronneberger2015u} for textual and visual explainability, respectively. Detailed explanations on baselines, evaluation metrics and training details are given in Appendix \ref{sec:appendixB}). Our proposed model can be utilized in a single task (keeping one task-specific private layers) or multi-task (keeping both visual and textual private layers) settings. In single task setting, there is no gating mechanism. 


\begin{table*}[hbt]

\caption{Results of proposed multitask model for textual and Visual Explainability, A1: Dot-product attention, A2: Multi-head attention,  CP-UNet: CLIP projection-based UNet, RW: Random weight, DC: Dice Coefficient, JS: Jaccard Similarity, mIOU: Mean Intersection over Union.}
\label{tab:res3}
\scalebox{0.80}{

\begin{tabular}{c|cccccccc|cccc}
\hline
\multirow{3}{*}{\textbf{Model}} &
  \multicolumn{8}{c|}{\textbf{Textual Explinability}} &
  \multicolumn{4}{c}{\textbf{Visual Explinability}} \\ \cline{2-13} 
 &
  \multicolumn{3}{c}{\textbf{ROUGE}} &
  \multicolumn{4}{c|}{\textbf{BLEU}} &
  \multirow{2}{*}{\textbf{HE}} &
  \multirow{2}{*}{\textbf{DC}} &
  \multirow{2}{*}{\textbf{JS}} &
  \multirow{2}{*}{\textbf{mIOU}} &
  \multirow{2}{*}{\textbf{HE}} \\ \cline{2-8}
 &
  \textbf{R1} &
  \textbf{R2} &
  \multicolumn{1}{c|}{\textbf{R-L}} &
  \textbf{B1} &
  \textbf{B2} &
  \textbf{B3} &
  \multicolumn{1}{c|}{\textbf{B4}} &
   &
   &
   &
   &
   \\ \hline
CP-UNet-T5\_A1 &
  60.94 &
  45.58 &
  \multicolumn{1}{c|}{60.43} &
  60.16 &
  53.32 &
  49.73 &
  \multicolumn{1}{c|}{46.93} &
  3.91 &
  \textbf{68.72} &
  54.72 &
  60.93 &
  \textbf{4.37} \\
CP-UNet-T5\_A1+RW &
  61.06 &
  46.33 &
  \multicolumn{1}{c|}{60.59} &
  60.63 &
  54.44 &
  51.05 &
  \multicolumn{1}{c|}{48.15} &
  4.07 &
  68.7 &
  \textbf{54.76} &
  61.29 &
  4.36 \\
CP-UNet-T5\_A2 &
  61.46 &
  45.63 &
  \multicolumn{1}{c|}{61.07} &
  60.86 &
  54.55 &
  50.93 &
  \multicolumn{1}{c|}{47.33} &
  4.31 &
  68.32 &
  54.11 &
  60.82 &
  4.28 \\
CP-UNet-T5\_A2+RW &
  61.67 &
  45.28 &
  \multicolumn{1}{c|}{61.21} &
  61.75 &
  55.24 &
  51.39 &
  \multicolumn{1}{c|}{47.82} &
  4.34 &
  68.38 &
  54.42 &
  59.93 &
  4.29 \\
CP-UNet-BART\_A1 &
  61.76 &
  45.68 &
  \multicolumn{1}{c|}{61.54} &
  61.68 &
  56.96 &
  52.26 &
  \multicolumn{1}{c|}{49.57} &
  4.38 &
  67.95 &
  53.67 &
  61.58 &
  4.25 \\
CP-UNet-BART\_A1+RW &
  63.06 &
  46.63 &
  \multicolumn{1}{c|}{62.57} &
  62.86 &
  56.55 &
  52.92 &
  \multicolumn{1}{c|}{49.33} &
  4.57 &
  67.32 &
  53.95 &
  61.13 &
  4.24 \\
CP-UNet-BART\_A2 &
  62.91 &
  46.93 &
  \multicolumn{1}{c|}{62.57} &
  62.44 &
  56.51 &
  53.03 &
  \multicolumn{1}{c|}{49.21} &
  4.42 &
  67.03 &
  53.69 &
  \textbf{62.53} &
  4.21 \\
CP-UNet-BART\_A2+RW &
  \textbf{63.54} &
  \textbf{47.36} &
  \multicolumn{1}{c|}{\textbf{63.07}} &
  \textbf{62.75} &
  \textbf{57.13} &
  \textbf{53.39} &
  \multicolumn{1}{c|}{\textbf{50.81}} &
  \textbf{4.59} &
  67.19 &
  53.03 &
  62.29 &
  4.23 \\ \hline
\end{tabular}%
}

\end{table*}

\begin{table}[hbt]
\caption{Results of different baselines and proposed Single task model for textual explainability}
\label{tab:res1}
\scalebox{0.52}{
\begin{tabular}{ccccccccc}
\hline
\multicolumn{1}{c|}{\multirow{2}{*}{\textbf{Model}}} & \multicolumn{3}{c|}{\textbf{ROUGE}}                 & \multicolumn{4}{c|}{\textbf{BLEU}}                          & \multirow{2}{*}{\textbf{HE}} \\ \cline{2-8}
\multicolumn{1}{c|}{}                       & \textbf{R1}    & \textbf{R2}    & \multicolumn{1}{c|}{\textbf{R-L}}   & \textbf{B1}    & \textbf{B2}    & \textbf{B3}    & \multicolumn{1}{c|}{\textbf{B4}}    &                     \\ \hline
\multicolumn{9}{c}{\textbf{Unimodal Baselines}}                                                                                                                              \\ \hline
\multicolumn{1}{c|}{T5-base}                & 59.97 & 44.01 & \multicolumn{1}{c|}{59.61} & 60.48 & 53.7  & 50.03 & \multicolumn{1}{c|}{47.14} & 3.67                \\
\multicolumn{1}{c|}{T5-large}               & 59.57 & 43.47 & \multicolumn{1}{c|}{59.07} & 58.87 & 52.43 & 48.83 & \multicolumn{1}{c|}{45.86} & 3.62                \\
\multicolumn{1}{c|}{Bart-base}              & 60.05 & 46.35 & \multicolumn{1}{c|}{59.86} & 60.55 & 56.46 & 50.52 & \multicolumn{1}{c|}{49.98} & 3.81                \\
\multicolumn{1}{c|}{Bart-large}             & 58.64 & 43.17 & \multicolumn{1}{c|}{58.15} & 58.4  & 51.62 & 47.95 & \multicolumn{1}{c|}{45.03} & 3.24                \\ \hline
\multicolumn{9}{c}{\textbf{Multimodal Baselines}}                                                                                                                            \\ \hline
\multicolumn{1}{c|}{VG-T5 (Dot-product)}   & 60.2  & 44.08 & \multicolumn{1}{c|}{59.7}  & 59.26 & 52.75 & 47.57 & \multicolumn{1}{c|}{46.52} & 3.85 \\
\multicolumn{1}{c|}{VG-T5 (Multi-head)}    & 60.85 & 44.97 & \multicolumn{1}{c|}{60.11} & 60.89 & 56.99 & 52.87 & \multicolumn{1}{c|}{49.29} & 3.93 \\
\multicolumn{1}{c|}{VG-BART (Dot-product)} & 60.84 & 45.76 & \multicolumn{1}{c|}{60.25} & 61.2  & 54.54 & 50.78 & \multicolumn{1}{c|}{47.81} & 3.91 \\
\multicolumn{1}{c|}{VG-BART (Multi-head)}  & 61.17 & 45.37 & \multicolumn{1}{c|}{60.8}  & 60.37 & 53.99 & 50.52 & \multicolumn{1}{c|}{47.57} & 4.26 \\ \hline
\multicolumn{9}{c}{\textbf{Proposed models}}                                                                                                                                 \\ \hline
\multicolumn{1}{c|}{CP-T5\_A1}              & 60.04 & 43.12 & \multicolumn{1}{c|}{59.32} & 59.55 & 52.87 & 49.02 & \multicolumn{1}{c|}{46.15} & 3.81                \\
\multicolumn{1}{c|}{CP-T5\_A1+RW}           & 60.15 & 43.56 & \multicolumn{1}{c|}{59.55} & 59.74 & 53.11 & 49.59 & \multicolumn{1}{c|}{46.72} & 3.83                \\
\multicolumn{1}{c|}{CP-T5\_A2}              & 61.16 & 44.69 & \multicolumn{1}{c|}{60.72} & 60.1  & 54.76 & 50.16 & \multicolumn{1}{c|}{47.31} & 4.21                \\
\multicolumn{1}{c|}{CP-T5\_A2+RW}           & 61.36 & 44.92 & \multicolumn{1}{c|}{60.97} & 60.59 & 54.34 & 50.88 & \multicolumn{1}{c|}{48.02} & 4.26                \\
\multicolumn{1}{c|}{CP-BART\_A1}            & 61.71 & 45.98 & \multicolumn{1}{c|}{61.27} & 62.17 & 55.55 & 51.73 & \multicolumn{1}{c|}{48.88} & 4.27                \\
\multicolumn{1}{c|}{CP-BART\_A1+RW}        & 62.37 & 46.51 & \multicolumn{1}{c|}{62.06} & 62.53 & 57.09 & 53.55 & \multicolumn{1}{c|}{50.85} & 4.32 \\
\multicolumn{1}{c|}{CP-BART\_A2}            & 61.99 & 46.11 & \multicolumn{1}{c|}{61.5}  & 62.43 & 55.72 & 51.96 & \multicolumn{1}{c|}{48.68} & 4.3                 \\
\multicolumn{1}{c|}{CP-BART\_A2+RW}        & \textbf{62.33} &\textbf{46.49} & \multicolumn{1}{c|}{\textbf{61.85}} & \textbf{62.44} & \textbf{55.9}  & \textbf{52.14} & \multicolumn{1}{c|}{\textbf{48.69}} & \textbf{4.39} \\ \hline
\end{tabular}%
}

\end{table}


\begin{table}[hbt]
\caption{Results of baselines and proposed Single task model for Visual explainability; V: Vision; L: Language; HE: Human Evaluation }
\label{tab:res2}
\scalebox{0.85}{
\begin{tabular}{ccccc}
\hline
\multicolumn{1}{c|}{\multirow{2}{*}{Model}} & \multicolumn{4}{c}{\textbf{Visual Explainability}}              \\ \cline{2-5} 
\multicolumn{1}{c|}{}                       & \textbf{DC}    & \textbf{JS}    &\textbf{mIOU}                            & \textbf{HE}   \\ \hline
\multicolumn{5}{c}{\textbf{Unimodal Baselines}                }                                            \\ \hline
\multicolumn{1}{c|}{DeepLabv3}              & 38.85 & 24.92 & 32.25                           & 1.79 \\
\multicolumn{1}{c|}{MobileNetV3}            & 39.49 & 25.49 & 32.16                           & 2.07 \\
\multicolumn{1}{c|}{FCN}                    & 39.21 & 25.29 & 31.97                           & 2.12 \\
\multicolumn{1}{c|}{UNet}                   & 41.89 & 27.35 & 31.79                           & 2.41 \\ \hline
\multicolumn{5}{c}{\textbf{Proposed Models}}                                                                \\ \hline
\multicolumn{1}{c|}{(V) CP-UNet}            & 65.71 & 51.86 & \textbf{63.03} & 3.83 \\
\multicolumn{1}{c|}{(V+L) CP-UNet}          & \textbf{66.22} & \textbf{52.45} & 62.95                           & \textbf{3.91} \\ \hline
\end{tabular}%
}

\end{table}

\subsection{Quantitative analysis}
We have conducted a statistical t-test on the results of our proposed model and other baselines and obtained a p-value less than 0.05.\\
{\bf (i) Single Task: Unimodal models}
The performance of unimodal models is detailed in Table~\ref{tab:res1} (textual explanations) and Table~\ref{tab:res2} (visual explanations). T5-base and BART-base models outperform their larger counterparts, possibly due to overfitting from excessive parameters given the limited dataset size (3222 instances).
For visual explanations, our CLIP-based UNet excels compared to baseline models using visual features from networks like ResNet, VGG19, AlexNet, etc., optimized for ImageNet, not memes. This superiority stems from CLIP's fine-tuning to better represent visual information through language supervision~\cite{radford2021learning}.

{\bf (ii) Single Task: Multimodal models} Our proposed multimodal models use dot product attention-based fusion (A1) and multi-head attention-based fusion (A2) techniques, combined with gated visual projection. According to the results (see~\ref{tab:res1} and~\ref{tab:res2}), our CLIP projection-based GPLMs outshine all other models. The top model, CP-BART\_A2 + RW, notably improves over previous best models by up to 2.28 ROUGE-1, 0.14 ROUGE-2, and 3.7 ROUGE-L scores. Using visual embeddings from CLIP with randomly initialized learnable weights (+RW) significantly enhances performance in textual explainability tasks. The language-aware CLIP-UNet model outperforms its unimodal counterpart, with improvements up to 1.01 in DC and 0.59 in JS scores, and substantial margins over the previous best unimodal model. However, the enhancement by the language-aware variant is marginal, likely because CLIP embeddings are optimized for visual rather than textual information.

{\bf (iii) Multi-Task: Shared-Private Architecture}
As evidenced by the results presented in Table ~\ref{tab:res3}, ~\ref{tab:res1}, and \ref{tab:res2}, it can be observed that the CLIP projection-based multimodal shared-private multitask approach outperforms all single task baselines by a significant margin, thus supporting the notion that training with multimodal explanations leads to enhanced performance in the generation of textual justifications and more precise identification of visual evidence. Notably, our most effective multitask model, CP-UNet-BART\_A2 + RW, which is optimized for text explanations, outperforms the best single-task textual explainability model (CP-BART) by 1.21 R1, 0.87 R2, and 1.22 R3. Additionally, the best multitasking model, CP-UNet-T5\_A1 + RW, which is optimized for visual explanations, outperforms the single task visual explainability model (CP\_UNet) by 2.48 DC, and 2.31 JS.

\textbf{(iv) Human Evaluation (HE):}
We conducted a human evaluation to assess the quality of generated explanations from our proposed methods. \textit{MExCCM} was evaluated based on the following criteria: 
{\bf 1 - Very Irrelevant: }
 The explanation does not address the topic or concept adequately. {\bf 5 - Very Relevant: }
The explanation is highly relevant to the topic or concept. Our analysis revealed notable findings regarding the relevance of different models in various settings. Specifically, when considering unimodal approaches, our best language-based model, CP-BART\_A2+RW, achieved an impressive average relevance score of 4.39 for textual explanations. On the other hand, our vision-based model, CP-UNet, obtained an average relevance score of 3.91 for visual explanations. Moving on to the multitask setting, our model CP-UNet-BART\_A1+RW demonstrated exceptional performance by achieving an average relevance score of 4.59 for textual explainability. Similarly, our model CP-UNet-T5\_A1 excelled in providing relevant visual explanations, securing an average relevance score of 4.37 ( refer to Table~\ref{tab:res3}, \ref{tab:res1} and \ref{tab:res2} for more details).

\subsection{Qualitative Analysis}
\begin{figure}[hbt]
	\centering
	\includegraphics[ angle= 90, origin=c, height = 11 cm, width = 8cm]{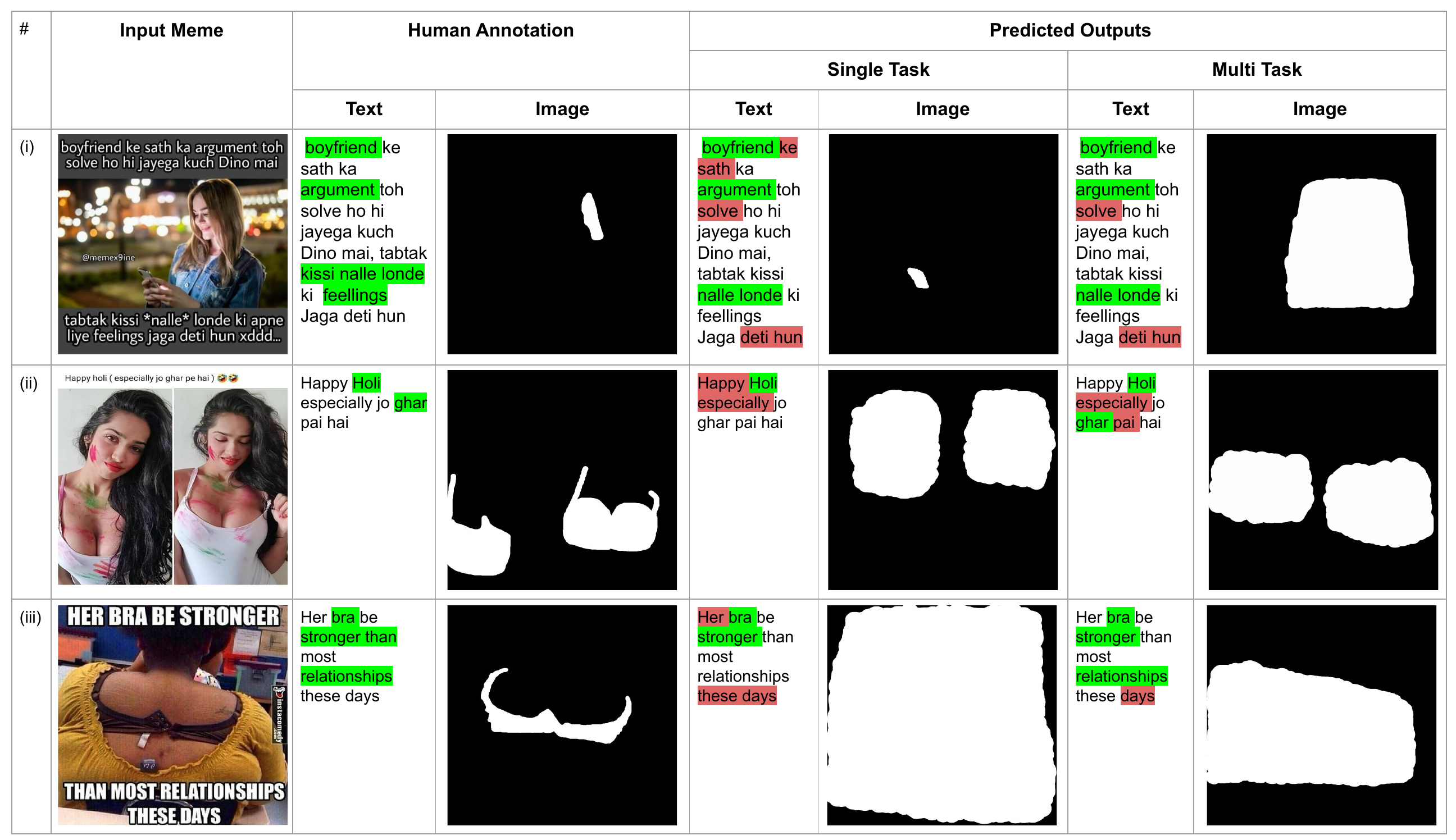}
	\caption {Human annotation vs. proposed model's visual and textual explanations; Green highlights indicate an agreement between the human annotator and the model. Red highlighted tokens are predicted by models, not by human annotators. }
	\label{fig:error}
\end{figure}

Figure~\ref{fig:error} presents results comparing visual and textual explainability of ground truth vs. model predictions.\\
(i) In the first meme, textual context ("boyfriend ke sath ka argument toh solve ho hi jayega...") is more vital than the visual. Both single-task and multi-task models identify the same number of correct rationales. The multi-task model better captures the visual aspect, representing the girl's face. (ii) In the second meme, visual cues surpass the textual message ("Happy Holi especially jo ghar pai hai"). The multi-task model more accurately identifies visual and textual cues than the single-task model. (iii) For the final meme, both modalities equally contribute to the meme's meaning. The multi-task model fares better, capturing most rationales with minor mistakes. Visually, the single-task model's prediction is less accurate than the multi-task model's.\\
From this qualitative analysis, we can conclude that (a) Multi-task-model is performing better than single task model, but visual explainability is still not convincing. More research is needed in this direction. (b) In cases where any one of the modalities dominates the others (example i and ii) single-task model performance is comparable to multitasking. (c) In cases where both modalities have an equal contribution, the multi-task model significantly performs better than the single-task model, which reveals that simultaneously learning both textual and visual explainability helps improve the performance of both tasks. 

\section{Conclusion and Future Work}
To encourage more research on explainable meme cyberbullying detection in code-mixed language, we introduced {\em MultiBully-Ex}, manually annotated with textual and visual explanations. This work introduces a CLIP projection-based multimodal shared-private architecture to generate rationales (textual explainability) and binary segmented image maps (visual explainability). Experimental results demonstrate that multitask models outperform the single-task model by a significant margin. We showed examples where visual modality is more effective than textual ones and vice versa, supporting the idea that multimodal explainable models provide better insight than unimodal approaches.  \\
Future attempts will be made to develop methods for identifying stereotypes in cyberbullying memes to capture implicit content and training models on a diverse dataset to enhance the performance of visual explainability.

\section{Limitation}
We have proposed a shared-private multimodal multitask architecture and a new benchmark dataset, {\em MultiBully-Ex}, to improve the explainability of cyberbullying memes in code-mixed Indian languages. However, there are some limitations to this approach: \\
1) Specifically, the textual explainability of memes is limited to the lexical level, which precludes the detection of implicit cyberbullying or stereotypes.\\
2) One of the main limitations of our work is its lack of generalizability to other code-mixed languages such as English and Spanish. However, this limitation can be addressed by fine-tuning the model on other code-mixed languages, which will enable it to capture the cultural nuances of the language.\\
3) Additionally, the visual explainability aspect of our approach, which involves predicting binary segmentation maps, is susceptible to the center bias commonly observed in computer vision models. This can impede the correct identification of visual cues that support the textual explanations, particularly for objects or features located in the corners or edges of the image.\\
4) This study is specifically dedicated to the analysis and understanding of memes in this image and text-based format. It is essential to highlight that our research delves into the unique characteristics and communication potential of static memes, distinct from the analysis of dynamic video memes. The latter, involving audiovisual elements, falls beyond the scope of our investigation.

\bibliography{anthology,custom}

\begin{thebibliography}{45}
\expandafter\ifx\csname natexlab\endcsname\relax\def\natexlab#1{#1}\fi

\bibitem[{Agrawal and Awekar(2018)}]{agrawal2018deep}
Sweta Agrawal and Amit Awekar. 2018.
\newblock Deep learning for detecting cyberbullying across multiple social media platforms.
\newblock In \emph{European conference on information retrieval}, pages 141--153. Springer.

\bibitem[{Ba et~al.(2016)Ba, Kiros, and Hinton}]{ba2016layer}
Jimmy~Lei Ba, Jamie~Ryan Kiros, and Geoffrey~E Hinton. 2016.
\newblock Layer normalization.
\newblock \emph{arXiv preprint arXiv:1607.06450}.

\bibitem[{Bengio et~al.(2009)Bengio, Louradour, Collobert, and Weston}]{bengio2009curriculum}
Yoshua Bengio, J{\'e}r{\^o}me Louradour, Ronan Collobert, and Jason Weston. 2009.
\newblock Curriculum learning.
\newblock In \emph{Proceedings of the 26th annual international conference on machine learning}, pages 41--48.

\bibitem[{Cao et~al.(2020)Cao, Lee, and Hoang}]{cao2020deephate}
Rui Cao, Roy Ka-Wei Lee, and Tuan-Anh Hoang. 2020.
\newblock Deephate: Hate speech detection via multi-faceted text representations.
\newblock In \emph{12th ACM conference on web science}, pages 11--20.

\bibitem[{Chan et~al.(2019)Chan, Cheung, and Wong}]{chan2019cyberbullying}
Tommy~KH Chan, Christy~MK Cheung, and Randy~YM Wong. 2019.
\newblock Cyberbullying on social networking sites: the crime opportunity and affordance perspectives.
\newblock \emph{Journal of Management Information Systems}, 36(2):574--609.

\bibitem[{Chen et~al.(2017)Chen, Papandreou, Schroff, and Adam}]{chen2017rethinking}
Liang-Chieh Chen, George Papandreou, Florian Schroff, and Hartwig Adam. 2017.
\newblock Rethinking atrous convolution for semantic image segmentation.
\newblock \emph{arXiv preprint arXiv:1706.05587}.

\bibitem[{Dadvar et~al.(2014)Dadvar, Trieschnigg, and Jong}]{dadvar2014experts}
Maral Dadvar, Dolf Trieschnigg, and Franciska~de Jong. 2014.
\newblock Experts and machines against bullies: A hybrid approach to detect cyberbullies.
\newblock In \emph{Canadian conference on artificial intelligence}, pages 275--281. Springer.

\bibitem[{Dice(1945)}]{dice1945measures}
Lee~R Dice. 1945.
\newblock Measures of the amount of ecologic association between species.
\newblock \emph{Ecology}, 26(3):297--302.

\bibitem[{Fleiss(1971)}]{fleiss1971measuring}
Joseph~L Fleiss. 1971.
\newblock Measuring nominal scale agreement among many raters.
\newblock \emph{Psychological bulletin}, 76(5):378.

\bibitem[{Garc{\'\i}a-Valero(2020)}]{garcia2020borreguero}
Benito Garc{\'\i}a-Valero. 2020.
\newblock Borreguero zuloaga, m. and vitacolonna, l.(eds.), the legacy of j{\'a}nos s. pet{\H{o}}fi. text linguistics, literary theory and semiotics.
\newblock \emph{Journal of Literary Semantics}, 49(1):61--64.

\bibitem[{Gomez et~al.(2020)Gomez, Gibert, Gomez, and Karatzas}]{gomez2020exploring}
Raul Gomez, Jaume Gibert, Lluis Gomez, and Dimosthenis Karatzas. 2020.
\newblock Exploring hate speech detection in multimodal publications.
\newblock In \emph{Proceedings of the IEEE/CVF Winter Conference on Applications of Computer Vision}, pages 1470--1478.

\bibitem[{He et~al.(2016)He, Zhang, Ren, and Sun}]{he2016deep}
Kaiming He, Xiangyu Zhang, Shaoqing Ren, and Jian Sun. 2016.
\newblock Deep residual learning for image recognition.
\newblock In \emph{Proceedings of the IEEE conference on computer vision and pattern recognition}, pages 770--778.

\bibitem[{Hee et~al.(2022)Hee, Lee, and Chong}]{DBLP:conf/www/HeeLC22}
Ming~Shan Hee, Roy~Ka{-}Wei Lee, and Wen{-}Haw Chong. 2022.
\newblock \href {https://doi.org/10.1145/3485447.3512260} {On explaining multimodal hateful meme detection models}.
\newblock In \emph{{WWW} '22: The {ACM} Web Conference 2022, Virtual Event, Lyon, France, April 25 - 29, 2022}, pages 3651--3655. {ACM}.

\bibitem[{Howard et~al.(2019)Howard, Sandler, Chu, Chen, Chen, Tan, Wang, Zhu, Pang, Vasudevan et~al.}]{howard2019searching}
Andrew Howard, Mark Sandler, Grace Chu, Liang-Chieh Chen, Bo~Chen, Mingxing Tan, Weijun Wang, Yukun Zhu, Ruoming Pang, Vijay Vasudevan, et~al. 2019.
\newblock Searching for mobilenetv3.
\newblock In \emph{Proceedings of the IEEE/CVF international conference on computer vision}, pages 1314--1324.

\bibitem[{Jaccard(1901)}]{jaccard1901distribution}
Paul Jaccard. 1901.
\newblock Distribution de la flore alpine dans le bassin des dranses et dans quelques r{\'e}gions voisines.
\newblock \emph{Bull Soc Vaudoise Sci Nat}, 37:241--272.

\bibitem[{Jha et~al.(2022)Jha, Dias, Lechervy, Moreno, Jangra, Pais, and Saha}]{jha2022combining}
Prince Jha, Ga{\"e}l Dias, Alexis Lechervy, Jose~G Moreno, Anubhav Jangra, Sebasti{\~a}o Pais, and Sriparna Saha. 2022.
\newblock Combining vision and language representations for patch-based identification of lexico-semantic relations.
\newblock In \emph{Proceedings of the 30th ACM International Conference on Multimedia}, pages 4406--4415.

\bibitem[{Karim et~al.(2021)Karim, Dey, Islam, Sarker, Menon, Hossain, Hossain, and Decker}]{karim2021deephateexplainer}
Md~Rezaul Karim, Sumon~Kanti Dey, Tanhim Islam, Sagor Sarker, Mehadi~Hasan Menon, Kabir Hossain, Md~Azam Hossain, and Stefan Decker. 2021.
\newblock Deephateexplainer: Explainable hate speech detection in under-resourced bengali language.
\newblock In \emph{2021 IEEE 8th International Conference on Data Science and Advanced Analytics (DSAA)}, pages 1--10. IEEE.

\bibitem[{Kiela et~al.(2020)Kiela, Firooz, Mohan, Goswami, Singh, Ringshia, and Testuggine}]{DBLP:conf/nips/KielaFMGSRT20}
Douwe Kiela, Hamed Firooz, Aravind Mohan, Vedanuj Goswami, Amanpreet Singh, Pratik Ringshia, and Davide Testuggine. 2020.
\newblock \href {https://proceedings.neurips.cc/paper/2020/hash/1b84c4cee2b8b3d823b30e2d604b1878-Abstract.html} {The hateful memes challenge: Detecting hate speech in multimodal memes}.
\newblock In \emph{Advances in Neural Information Processing Systems 33: Annual Conference on Neural Information Processing Systems 2020, NeurIPS 2020, December 6-12, 2020, virtual}.

\bibitem[{Lewis et~al.(2019)Lewis, Liu, Goyal, Ghazvininejad, Mohamed, Levy, Stoyanov, and Zettlemoyer}]{lewis2019bart}
Mike Lewis, Yinhan Liu, Naman Goyal, Marjan Ghazvininejad, Abdelrahman Mohamed, Omer Levy, Ves Stoyanov, and Luke Zettlemoyer. 2019.
\newblock Bart: Denoising sequence-to-sequence pre-training for natural language generation, translation, and comprehension.
\newblock \emph{arXiv preprint arXiv:1910.13461}.

\bibitem[{Lin and Och(2004)}]{lin2004automatic}
Chin-Yew Lin and Franz~Josef Och. 2004.
\newblock Automatic evaluation of machine translation quality using longest common subsequence and skip-bigram statistics.
\newblock In \emph{Proceedings of the 42nd Annual Meeting of the Association for Computational Linguistics (ACL-04)}, pages 605--612.

\bibitem[{Liu et~al.(2016)Liu, Qiu, and Huang}]{liu2016recurrent}
Pengfei Liu, Xipeng Qiu, and Xuanjing Huang. 2016.
\newblock Recurrent neural network for text classification with multi-task learning.
\newblock \emph{arXiv preprint arXiv:1605.05101}.

\bibitem[{Long et~al.(2015)Long, Shelhamer, and Darrell}]{long2015fully}
Jonathan Long, Evan Shelhamer, and Trevor Darrell. 2015.
\newblock Fully convolutional networks for semantic segmentation.
\newblock In \emph{Proceedings of the IEEE conference on computer vision and pattern recognition}, pages 3431--3440.

\bibitem[{Lu et~al.(2018)Lu, Neves, Carvalho, Zhang, and Ji}]{lu2018visual}
Di~Lu, Leonardo Neves, Vitor Carvalho, Ning Zhang, and Heng Ji. 2018.
\newblock Visual attention model for name tagging in multimodal social media.
\newblock In \emph{Proceedings of the 56th Annual Meeting of the Association for Computational Linguistics (Volume 1: Long Papers)}, pages 1990--1999.

\bibitem[{Lundberg and Lee(2017)}]{lundberg2017unified}
Scott~M Lundberg and Su-In Lee. 2017.
\newblock A unified approach to interpreting model predictions.
\newblock \emph{Advances in neural information processing systems}, 30.

\bibitem[{Maity et~al.(2022{\natexlab{a}})Maity, Jha, Saha, and Bhattacharyya}]{DBLP:conf/sigir/MaityJ0B22}
Krishanu Maity, Prince Jha, Sriparna Saha, and Pushpak Bhattacharyya. 2022{\natexlab{a}}.
\newblock \href {https://doi.org/10.1145/3477495.3531925} {A multitask framework for sentiment, emotion and sarcasm aware cyberbullying detection from multi-modal code-mixed memes}.
\newblock In \emph{{SIGIR} '22: The 45th International {ACM} {SIGIR} Conference on Research and Development in Information Retrieval, Madrid, Spain, July 11 - 15, 2022}, pages 1739--1749. {ACM}.

\bibitem[{Maity et~al.(2022{\natexlab{b}})Maity, Jha, Saha, and Bhattacharyya}]{maity2022multitask}
Krishanu Maity, Prince Jha, Sriparna Saha, and Pushpak Bhattacharyya. 2022{\natexlab{b}}.
\newblock A multitask framework for sentiment, emotion and sarcasm aware cyberbullying detection from multi-modal code-mixed memes.

\bibitem[{Mathew et~al.(2020)Mathew, Saha, Yimam, Biemann, Goyal, and Mukherjee}]{mathew2020hatexplain}
Binny Mathew, Punyajoy Saha, Seid~Muhie Yimam, Chris Biemann, Pawan Goyal, and Animesh Mukherjee. 2020.
\newblock Hatexplain: A benchmark dataset for explainable hate speech detection.
\newblock \emph{arXiv preprint arXiv:2012.10289}.

\bibitem[{Myers-Scotton(1997)}]{cm1}
Carol Myers-Scotton. 1997.
\newblock \emph{Duelling languages: Grammatical structure in codeswitching}.
\newblock Oxford University Press.

\bibitem[{Papineni et~al.(2002)Papineni, Roukos, Ward, and Zhu}]{papineni2002bleu}
Kishore Papineni, Salim Roukos, Todd Ward, and Wei-Jing Zhu. 2002.
\newblock Bleu: a method for automatic evaluation of machine translation.
\newblock In \emph{Proceedings of the 40th annual meeting of the Association for Computational Linguistics}, pages 311--318.

\bibitem[{Paul and Saha(2020)}]{paul2020cyberbert}
Sayanta Paul and Sriparna Saha. 2020.
\newblock Cyberbert: Bert for cyberbullying identification.
\newblock \emph{Multimedia Systems}, pages 1--8.

\bibitem[{Pramanick et~al.(2021)Pramanick, Sharma, Dimitrov, Akhtar, Nakov, and Chakraborty}]{DBLP:conf/emnlp/PramanickSDAN021}
Shraman Pramanick, Shivam Sharma, Dimitar Dimitrov, Md.~Shad Akhtar, Preslav Nakov, and Tanmoy Chakraborty. 2021.
\newblock \href {https://doi.org/10.18653/v1/2021.findings-emnlp.379} {{MOMENTA:} {A} multimodal framework for detecting harmful memes and their targets}.
\newblock In \emph{Findings of the Association for Computational Linguistics: {EMNLP} 2021, Virtual Event / Punta Cana, Dominican Republic, 16-20 November, 2021}, pages 4439--4455. Association for Computational Linguistics.

\bibitem[{Radford et~al.(2021)Radford, Kim, Hallacy, Ramesh, Goh, Agarwal, Sastry, Askell, Mishkin, Clark et~al.}]{radford2021learning}
Alec Radford, Jong~Wook Kim, Chris Hallacy, Aditya Ramesh, Gabriel Goh, Sandhini Agarwal, Girish Sastry, Amanda Askell, Pamela Mishkin, Jack Clark, et~al. 2021.
\newblock Learning transferable visual models from natural language supervision.
\newblock In \emph{International Conference on Machine Learning}, pages 8748--8763. PMLR.

\bibitem[{Raffel et~al.(2020)Raffel, Shazeer, Roberts, Lee, Narang, Matena, Zhou, Li, Liu et~al.}]{raffel2020exploring}
Colin Raffel, Noam Shazeer, Adam Roberts, Katherine Lee, Sharan Narang, Michael Matena, Yanqi Zhou, Wei Li, Peter~J Liu, et~al. 2020.
\newblock Exploring the limits of transfer learning with a unified text-to-text transformer.
\newblock \emph{J. Mach. Learn. Res.}, 21(140):1--67.

\bibitem[{Ribeiro et~al.(2016)Ribeiro, Singh, and Guestrin}]{ribeiro2016should}
Marco~Tulio Ribeiro, Sameer Singh, and Carlos Guestrin. 2016.
\newblock " why should i trust you?" explaining the predictions of any classifier.
\newblock In \emph{Proceedings of the 22nd ACM SIGKDD international conference on knowledge discovery and data mining}, pages 1135--1144.

\bibitem[{Rijhwani et~al.(2017)Rijhwani, Sequiera, Choudhury, Bali, and Maddila}]{rijhwani2017estimating}
Shruti Rijhwani, Royal Sequiera, Monojit Choudhury, Kalika Bali, and Chandra~Shekhar Maddila. 2017.
\newblock Estimating code-switching on twitter with a novel generalized word-level language detection technique.
\newblock In \emph{Proceedings of the 55th annual meeting of the association for computational linguistics (volume 1: long papers)}, pages 1971--1982.

\bibitem[{Ronneberger et~al.(2015)Ronneberger, Fischer, and Brox}]{ronneberger2015u}
Olaf Ronneberger, Philipp Fischer, and Thomas Brox. 2015.
\newblock U-net: Convolutional networks for biomedical image segmentation.
\newblock In \emph{International Conference on Medical image computing and computer-assisted intervention}, pages 234--241. Springer.

\bibitem[{ROUGE(2004)}]{rouge2004package}
Lin~CY ROUGE. 2004.
\newblock A package for automatic evaluation of summaries.
\newblock In \emph{Proceedings of Workshop on Text Summarization of ACL, Spain}.

\bibitem[{Smith et~al.(2008)Smith, Mahdavi, Carvalho, Fisher, Russell, and Tippett}]{smith2008cyberbullying}
Peter~K Smith, Jess Mahdavi, Manuel Carvalho, Sonja Fisher, Shanette Russell, and Neil Tippett. 2008.
\newblock Cyberbullying: Its nature and impact in secondary school pupils.
\newblock \emph{Journal of child psychology and psychiatry}, 49(4):376--385.

\bibitem[{Sticca et~al.(2013)Sticca, Ruggieri, Alsaker, and Perren}]{sticca2013longitudinal}
Fabio Sticca, Sabrina Ruggieri, Fran{\c{c}}oise Alsaker, and Sonja Perren. 2013.
\newblock Longitudinal risk factors for cyberbullying in adolescence.
\newblock \emph{Journal of community \& applied social psychology}, 23(1):52--67.

\bibitem[{Suryawanshi et~al.(2020)Suryawanshi, Chakravarthi, Arcan, and Buitelaar}]{suryawanshi2020multimodal}
Shardul Suryawanshi, Bharathi~Raja Chakravarthi, Mihael Arcan, and Paul Buitelaar. 2020.
\newblock Multimodal meme dataset (multioff) for identifying offensive content in image and text.
\newblock In \emph{Proceedings of the Second Workshop on Trolling, Aggression and Cyberbullying}, pages 32--41.

\bibitem[{Tsai et~al.(2019)Tsai, Bai, Liang, Kolter, Morency, and Salakhutdinov}]{tsai2019multimodal}
Yao-Hung~Hubert Tsai, Shaojie Bai, Paul~Pu Liang, J~Zico Kolter, Louis-Philippe Morency, and Ruslan Salakhutdinov. 2019.
\newblock Multimodal transformer for unaligned multimodal language sequences.
\newblock In \emph{Proceedings of the conference. Association for Computational Linguistics. Meeting}, volume 2019, page 6558. NIH Public Access.

\bibitem[{Ybarra et~al.(2006)Ybarra, Mitchell, Wolak, and Finkelhor}]{ybarra2006examining}
Michele~L Ybarra, Kimberly~J Mitchell, Janis Wolak, and David Finkelhor. 2006.
\newblock Examining characteristics and associated distress related to internet harassment: findings from the second youth internet safety survey.
\newblock \emph{Pediatrics}, 118(4):e1169--e1177.

\bibitem[{Yu et~al.(2021)Yu, Dai, Liu, and Fung}]{yu2021vision}
Tiezheng Yu, Wenliang Dai, Zihan Liu, and Pascale Fung. 2021.
\newblock Vision guided generative pre-trained language models for multimodal abstractive summarization.
\newblock \emph{arXiv preprint arXiv:2109.02401}.

\bibitem[{Zaidan et~al.(2007)Zaidan, Eisner, and Piatko}]{zaidan2007using}
Omar Zaidan, Jason Eisner, and Christine Piatko. 2007.
\newblock Using “annotator rationales” to improve machine learning for text categorization.
\newblock In \emph{Human language technologies 2007: The conference of the North American chapter of the association for computational linguistics; proceedings of the main conference}, pages 260--267.

\bibitem[{Zhang et~al.(2018)Zhang, Fu, Liu, and Huang}]{zhang2018adaptive}
Qi~Zhang, Jinlan Fu, Xiaoyu Liu, and Xuanjing Huang. 2018.
\newblock Adaptive co-attention network for named entity recognition in tweets.
\newblock In \emph{Thirty-Second AAAI Conference on Artificial Intelligence}.

\end{thebibliography}
\bibliographystyle{acl_natbib}

\appendix

\appendix

\section{Attention Mechanism}
\label{sec:A}
\subsection{Dot Product Attention Based Fusion}
In this type of fusion mechanism, we begin by projecting the visual features to the same dimensional space as the textual features (Eq. \ref{eq4}). Then, the dot-product is calculated, and the softmax function is applied (Eq. \ref{eq5}). Finally, the input textual features are combined with the attention-weighted visual features and projected through a linear transformation to generate the vision-guided textual features (Eq. \ref{eq6}).
\begin{equation}
\label{eq4}
Z^{'}_{v}=Z_{v}W_{1}
\end{equation}
\begin{equation}
\label{eq5}
A=Softmax(Z_{t}Z^{'}_{v})
\end{equation}
\begin{equation}
\label{eq6}
Z^{'}_{t}=Concat(Z_{t},AZ_{v})W_{2}
\end{equation}
\subsection{Multi-head Attention Based Fusion}
In this type of fusion mechanism, a multi-head attention mechanism based on vision guidance is used for text-vision fusion. Query, Key, and Value are all projected linearly from the input text and visual components (Eq. \ref{eq7} - Eq. \ref{eq9}). Cross-modal attention is utilized to gather the text-queried visual features (Eq. \ref{eq10}). Finally, the final output representation is created by combining input textual features and text-queried visual features (Eq. \ref{eq11}).
\begin{equation}
\label{eq7}
Q=Z_{t}W_{q}
\end{equation}
\begin{equation}
\label{eq8}
K=Z_{v}W_{k}
\end{equation}
\begin{equation}
\label{eq9}
V=Z_{v}W_{v}
\end{equation}
\begin{equation}
\label{eq10}
O=CMA(Q,K,V)
\end{equation}
\begin{equation}
\label{eq11}
Z^{'}_{t}= Concat(Z_{t},O)W_{3}
\end{equation}

\section{Experimental Setups}
\label{sec:appendixB}
\subsection{Generation Baselines}
\textbf{BART \cite{lewis2019bart}:} BART is an encoder-decoder-based transformer model which is mainly pre-
trained for text generation tasks such as summarization and translation. BART is pre-
trained with various denoising pretraining objectives such as token masking, sentence
permutation, sentence rotation etc.
\\
\textbf{T5 \cite{raffel2020exploring}: }T5 is also an encoder-decoder-based transformer model which aims to solve all
the text-to-text generation problems. The main difference between BART and T5 is the
pre-training objective. In T5, the transformer is pre-trained with a denoising objective
where 15
these masked tokens whereas, during pre-training of BART, the decoder generates the
complete input sequence\\
\textbf{VG-BART \cite{yu2021vision}:} VG-BART is a multimodal variant of BART proposed by \citet{yu2021vision} that uses a text-vision fusion mechanism inside BART encoder.\\
\textbf{VG-T5 \cite{yu2021vision}: }The work of \citet{yu2021vision} presents VG-T5, a multimodal version of T5 which incorporates a text-visual fusion technique within the T5 encoder.
\subsection{Segmentation Baselines}
\textbf{Fully Convolutional Network (FCN): }FCN~\cite{long2015fully} is a type of CNN that can segment images of any size, it was one of the first models that can handle variable size inputs, now it is a standard in most segmentation models. The model upsamples the feature maps from lower layers and combine them with higher layer feature maps to produce the final segmentation mask.\\
\textbf{DeepLabv3}
DeepLabv3~\cite{chen2017rethinking}, developed by Google in 2017, is a state-of-the-art semantic image segmentation model that utilizes an encoder-decoder architecture incorporating atrous convolution and skip connections to enhance segmentation accuracy.
\\
\textbf{MobileNetv3: }
MobileNetv3~\cite{howard2019searching} is a lightweight neural network architecture utilizes a combination of depthwise convolution and bottlenecks blocks to achieve high efficiency and accuracy. It also uses a new neural architecture search method to find the optimal combination of building blocks.\\
\textbf{UNet: }UNet~\cite{ronneberger2015u} is a convolutional neural network, utilizes a "U" shaped architecture that combines the feature information from a downsampling path with the upsampled output from an upsampling path. The architecture also uses skip connections to concatenate the feature maps from the downsampling path to the upsampling path, which helps to improve segmentation performance. 
\subsection{Evaluation Metrics}
We present the scores of five automated evaluation metrics, including ROUGE~\citep{rouge2004package} and BLEU~\citep{papineni2002bleu}, which are used to measure the performance of the textual explainability, as well as Dice Coefficient (DC)~\cite{dice1945measures}, Jaccard Similarity (JS)~\cite{jaccard1901distribution}, and mean Intersection over Union (mIOU), which are used to evaluate the visual explainability.

(i) {\bf BLEU:} One of the earliest metrics to be used to measure the similarity between two phrases is BLEU. It was first proposed for machine translation and is described as the geometric mean of n-gram precision scores times a brevity penalty for short sentences. We apply the smoothed BLEU in our experiments as defined in~\citep{lin2004automatic}.

(ii) {\bf ROUGE-L:} ROUGE was first presented for the assessment of summarization systems, and this evaluation is carried out by comparing overlapping n-grams, word sequences, and word pairs. In this work, we employ ROUGE-1 (unigram), ROUGE-2 (bigram) and ROUGE-L version, which measures the longest common subsequences between a pair of phrases.

(iii) {\bf Dice Coefficient:}
The Dice coefficient is a similarity metric used in image segmentation to measure the similarity between two sets. It ranges from 0 to 1, where 1 indicates perfect match and 0 indicates no match. The formula for Dice coefficient is $(2 * |A \cap B|) / (|A| + |B|)$, where A and B are the two sets being compared. It is particularly useful when working with imbalanced datasets.

(iv) {\bf Jaccard Similarity:}
Jaccard similarity is a similarity metric used to measure the similarity between two sets, it is often used in natural language processing, information retrieval and image segmentation. It ranges from 0 to 1, where 1 indicates perfect match and 0 indicates no match. The formula for Jaccard similarity is $|A \cap B| / |A \cup B|$, where A and B are the two sets being compared. 

(v) {\bf mIOU:}
Mean Intersection over Union (mIOU) is an evaluation metric used in image segmentation tasks, it is the mean of the Intersection over Union (IoU) scores for all the classes. It is used to measure the similarity of predicted segmentation maps with ground truth segmentation maps, unlike Jaccard similarity which is used to measure the similarity between two sets.
\subsection{Training Details}
In this section, we detail various hyperparameters and
experimental settings used in our work.
We have performed all the experiments on Tyrone machine with Intel’s Xeon W-2155 Processor having 196 Gb DDR4 RAM and 11 Gb Nvidia 1080Ti GPU. We have randomly chosen 70\% of the data for training, 10\% for validation, and the remaining 20\% for testing. We have executed all of the models five times, and the average results have been reported. We have used BART ~\cite{lewis2019bart}, T5~\cite{raffel2020exploring} as the base model for our proposed model. All the models are trained for a maximum of 40 epochs and a batch size of 32. Adam optimizer is used to train the model with an epsilon value of 0.00000001. All the models are implemented using Scikit-Learn\footnote{\url{https://scikit-learn.org/stable/}} and pytorch\footnote{\url{https://pytorch.org/}} as a backend.
\label{sec:B}

\section{Annotations}
\subsection{Annotation Guidelines}
\label{sec:ann-guide}
We follow cyberbullying definition by \cite{smith2008cyberbullying} for our annotation process. In order to help and guide our annotators, we provide them with several examples of memes with textual and visual explanations marked. We first read the entire text present inside the memes for rationale annotations and looked at the depicted visual clues. Each lexicon was marked either Bully or Non-bully based on the visual and textual context. Additionally, visual regions were segmented that prominently justified the rationale annotations for visual explanations.

\subsection{Daywise Schedule}
\label{sec:days}

\begin{itemize}
 \item \textbf{Day 1 and Day 4: } Each annotator was assigned to annotate rationales for 150 memes. They were instructed to annotate 30 memes per batch within one hour, followed by a mandatory break of 10 minutes (cf.  Section~\ref{sec: ethics}).
\item \textbf{Day  2 and Day 5: } Each annotator was assigned to highlight the visual regions that could justify the rationale annotations. 
\item \textbf{Day 3: } We arrange meetings with the annotators to ensure that their mental well-being is not adversely affected during the annotation process (cf. Section~\ref{sec: ethics}).
\end{itemize}

\subsection{Annotation cost}
\label{sec:cost}
The process of annotating multimodal explanation is time-consuming and expensive, with each meme sample requiring 2-3 minutes for textual and visual explanation each. We initially hired 10 annotators and selected 3 best annotators among them. An honorarium of 5 INR was offered per sample due to the inherent complexity, which was ensured to be appropriate considering the 160-750 INR minimum wage/day based on the Minimum Wages Act, 1948\footnote{\url{https://en.wikipedia.org/wiki/List_of_countries_by_minimum_wage}
} in India (where the annotations were done) based on the average number of annotations across all annotators per day. The entire annotation process took approximately 10 weeks to complete following daywise schedule.

\textbf{Ethics note: }
Repetitive consumption of online abuse could distress mental health conditions \cite{ybarra2006examining}. Therefore, we advised annotators to take periodic breaks and not do the annotations in one sitting. Besides, we had weekly meetings with them to ensure the annotations did not have any adverse effect on their mental health.
\label{sec: ethics}
\subsection{Statistics of Annotated Multimodal Explanations}
\label{sec:stats}
\begin{figure}[hbt]
	\centering
	\includegraphics[height = 10 cm, width = 8cm]{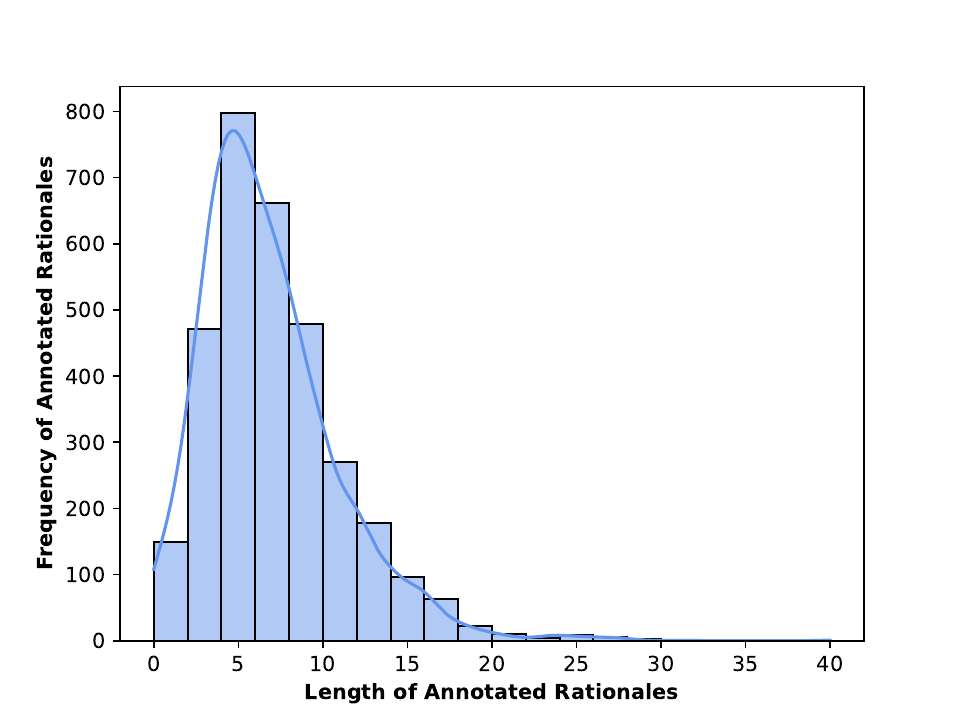}
 \caption{Distribution for Length of Meme Text}
	\label{fig:memetext}
\end{figure}

\begin{figure}[hbt]
	\centering
	\includegraphics[height = 10 cm, width = 8cm]{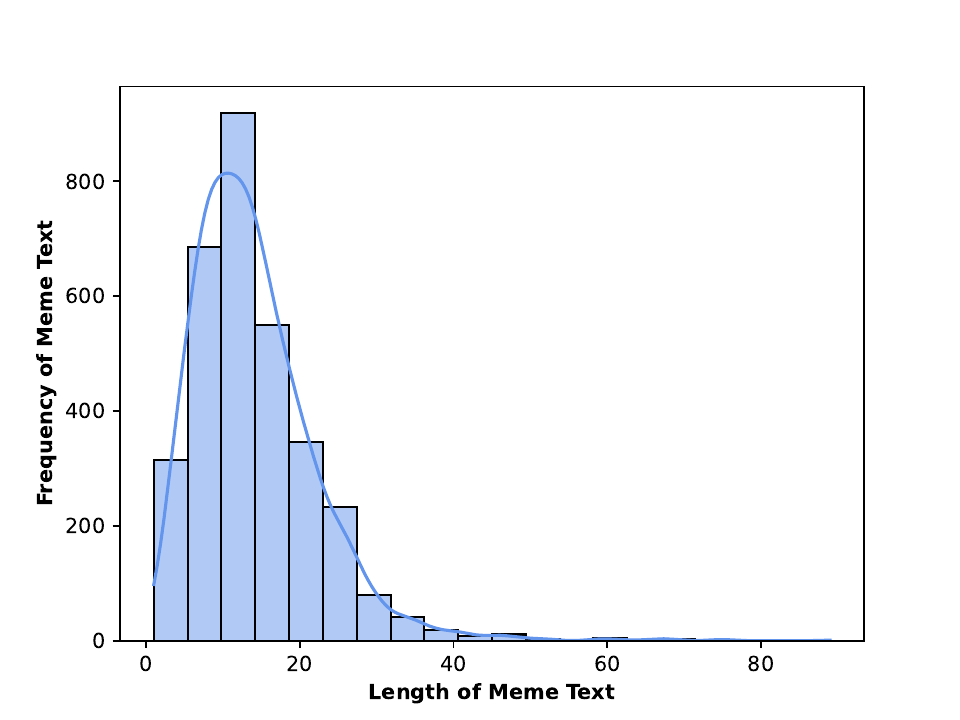}
 \caption{Distribution for Length of Annotated Rationales}
	\label{fig:rationale}
\end{figure}

\begin{figure}[hbt]
	\centering
	\includegraphics[height = 10 cm, width = 8cm]{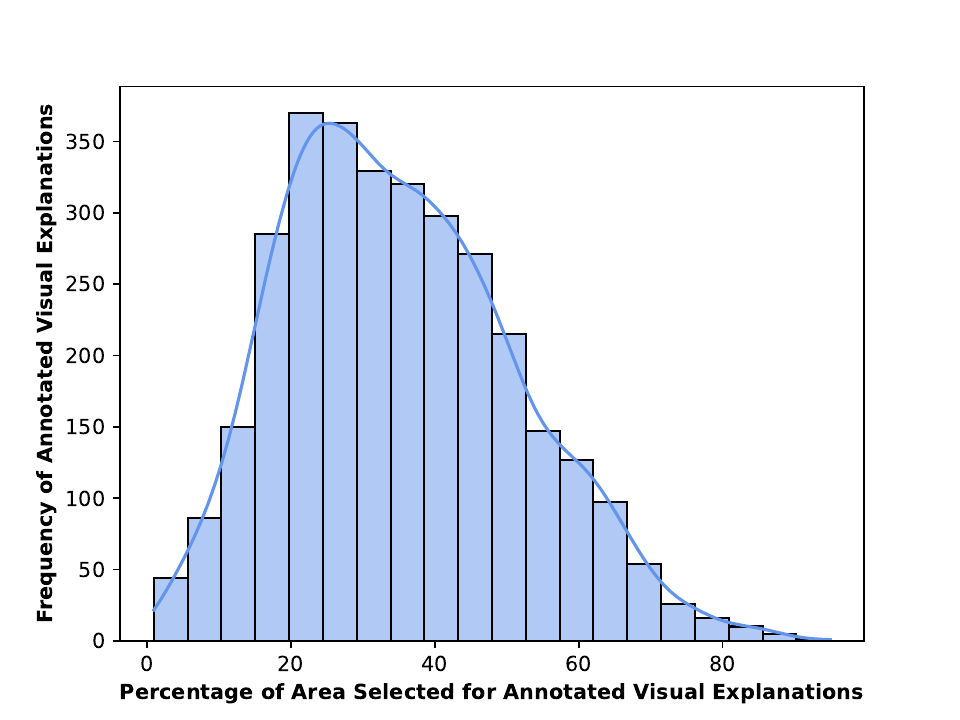}
 \caption{Distribution for Annotated Visual Explanations}
	\label{fig:area}
\end{figure}

Figure~\ref{fig:memetext} illustrates the distribution of meme text. The figure showcases that the length of meme text typically falls within the range of 0 to 80 characters. Upon conducting calculations, the average length of meme text was determined to be approximately 14.12 characters. In a similar vein, the length of rationales ranges from 0 to 40, as depicted in Figure~\ref{fig:rationale}. The average token length of annotated rationales was observed to be around 6.79. Furthermore, we observed that, on average, 35.18\% of the image area is dedicated to visual explanations for cyberbullying memes. The distribution for the percentage of area selected for annotated visual explanations can be found in Figure~\ref{fig:area}.

\end{document}